\documentclass[9pt,technote,compsoc]{IEEEtran}
\usepackage{amssymb}  

\ifCLASSOPTIONcompsoc
  \usepackage[nocompress]{cite}
\else
  \usepackage{cite}
\fi
\ifCLASSINFOpdf
  \usepackage[pdftex]{graphicx}
  \graphicspath{{./}}
  \DeclareGraphicsExtensions{.pdf,.jpeg,.png}
\else
\fi
\usepackage[cmex10]{amsmath}
\usepackage{algorithmic}

\usepackage{hyperref}

\begin{document}
\bstctlcite{IEEEexample:BSTcontrol}
%
\title{Multimodal Multipart Learning\\for Action Recognition in Depth Videos}
%
%
%
%

\author{Amir~Shahroudy,~\IEEEmembership{Student Member,~IEEE,}
        Gang~Wang,~\IEEEmembership{Member,~IEEE,}
        Tian-Tsong~Ng,~\IEEEmembership{Member,~IEEE,}
        and~Qingxiong~Yang,~\IEEEmembership{Member,~IEEE}
\IEEEcompsocitemizethanks{\IEEEcompsocthanksitem A. Shahroudy and G. Wang are with the School of Electrical and Electronic Engineering, Nanyang Technological University, Singapore, 639798.\protect\\
E-mail: \{amir3,wanggang\}@ntu.edu.sg.
\IEEEcompsocthanksitem A. Shahroudy and T.-T. Ng are with the Institute for Infocomm Research, 1 Fusionopolis Way, Singapore, 138632.\protect\\
E-mail: \{stusam,ttng\}@i2r.a-star.edu.sg.
\IEEEcompsocthanksitem Q. Yang is with the Department of Computer Science, City University of Hong Kong, Hong Kong.\protect\\
E-mail: qiyang@cityu.edu.hk.}
}

%
%

\markboth{}%
{}
%



\IEEEtitleabstractindextext{%
\begin{abstract}
\makebox[212pt][s]{The articulated and complex nature of human actions makes} \par
\makebox[252.5pt][s]{the task of action recognition difficult. One approach to handle this} \par
\makebox[252.5pt][s]{complexity is dividing it to the kinetics of body parts and analyzing} \par
\makebox[252.5pt][s]{the actions based on these partial descriptors. We propose a joint} \par
\makebox[252.5pt][s]{sparse regression based learning method which utilizes the structured} \par 
\makebox[252.5pt][s]{sparsity to model each action as a combination of multimodal features} \par
\makebox[252.5pt][s]{from a sparse set of body parts. To represent dynamics and appearance} \par 
\makebox[252.5pt][s]{of parts, we employ a heterogeneous set of depth and skeleton based} \par
\makebox[252.5pt][s]{features. The proper structure of multimodal multipart features are} \par
\makebox[252.5pt][s]{formulated into the learning framework via the proposed hierarchical} \par
\makebox[252.5pt][s]{mixed norm, to regularize the structured features of each part and to} \par
\makebox[252.5pt][s]{apply sparsity between them, in favor of a group feature selection.} \par
\makebox[252.5pt][s]{Our experimental results expose the effectiveness of the proposed} \par
\makebox[252.5pt][s]{learning method in which it outperforms other methods in all three tested} \par
datasets while saturating one of them by achieving perfect accuracy.
\end{abstract}

\begin{IEEEkeywords}
\makebox[197.8pt][s]{Action recognition, Kinect, Joint sparse regression,} \par  
Mixed norms, Structured sparsity, Group feature selection
\end{IEEEkeywords}}

\maketitle
\IEEEdisplaynontitleabstractindextext

%
\IEEEpeerreviewmaketitle
\vspace*{-0.7cm}
\section{Introduction}
\IEEEPARstart{H}{uman} actions consist of simultaneous flow of different body parts. Based on this complex articulated essence of human movements, the analysis of these signals could be highly complicated. To ease the task of classification, actions could be broken down into their components. This is done by a body part detection on depth sequences of human body movements \cite{shotton2011CVPR}. Having the 3D locations of body joints in the scene, we can separate the complicated motion of body into a concurrent set of behaviors on major skeleton joints; therefore human action sequences could be considered as multipart signals. Throughout this paper, we use the term ``part'' to denote each body joint as defined in \cite{shotton2011CVPR}.

Limiting the learning into skeleton based features cannot deliver high levels of performance in action recognition, because: (1) most of the usual human actions are defined based on the interaction of body with other objects, and (2) depth based skeleton data is not always accurate due to the noise and occlusion of body parts. To alleviate these issues, different depth based appearance features can be leveraged. The work in \cite{actionletPAMI} proposed LOP (local occupancy patterns) around each of the body joints in order to represent 3D appearance of the interacting objects. Another solution is HON4D (histogram of oriented 4D normals) \cite{HON4D}, which gives more descriptive and robust models of the local depth based appearance and motion, around the joints. Based on the complementary properties of mentioned features, it is beneficial to utilize all of them as different descriptors for each joint. Combining heterogeneous features of each part of the skeleton, leads into a multimodal-multipart combination, which demands sophisticated fusion algorithms.

An interesting approach to handle the articulation of actions was recently proposed by \cite{actionletPAMI}. As the key intuition, they have shown each individual action class can be represented by the behavior and appearance of few informative joints in the body. They utilized a data mining technique to find these discriminative sets of joints for each class of the available actions and tied up the features of those parts as ``actionlets''. They employed a multi-kernel learning method to build up ensembles of actionlets as kernels for action classification. This method is highly robust against the noise in depth maps, and the results show its strength to characterize the human body motion and also human-object interactions. However the downside of this approach is the inconsistency of their heuristic selection process (mining actionlets) with the following learning step. Moreover, it simply concatenates different types of features for multimodal fusion, which is another drawback of this work. In this fashion, achieving the optimal combination of features regarding the classification task cannot be guaranteed.

To overcome the limitations mentioned above, we propose a joint structured sparsity regression based learning method which integrates part selection into the learning process considering the heterogeneity of features for each joint. We associate all the features for each part as a bundle and apply a group sparsity regularization to select a small number of active parts for each action class. To model the precise hierarchy of the multimodal-multipart features in an integrated learning and selection framework, we propose a hierarchical mixed norm which includes three levels of regularization over learning weights. To apply the modality based coupling over heterogeneous features of each part, it applies a mixed norm with two degrees of ``diversity'' induction \cite{738251}, followed by a group sparsity among the feature groups of different parts to apply part selection.

The main contributions of this paper are two\--fold: First, we integrated the part selection process into our learning in order to select discriminative body parts for different action classes latently, and utilize them to learn classifiers. Second, a hierarchical mixed norm is proposed to apply the desired simultaneous sparsity and regularization over different levels of learning weights corresponding to our special multimodal-multipart features in a joint group sparsity regression framework.

We evaluate our method on three challenging depth based action recognition datasets: MSR\--DailyActivity dataset~\cite{actionletPAMI}, MSR\--Action3D dataset \cite{msraction3ddataset}, and 3D\--ActionPairs dataset \cite{HON4D}. Our experimental results show that the proposed method is superior to other available methods for action recognition on depth sequences.

The rest of this paper is organized as follows: Section 2  reviews the related works on depth based action recognition, joint sparse regression, mixed norms, and multitask learning. Section 3 presents the proposed integrated feature selection and learning scheme. It also introduces the new multimodal-multi part mixed norm which applies regularization and group sparsity into the proposed learning model. Experimental results on three above-mentioned benchmarks are covered in section 4 and we conclude the paper in section 5.

\section{Related Work}

Visual features extracted from depth signals can be classified into two major classes. The first are skeleton based features, which extract information from the provided 3D locations of body joints on each frame of the sequence. Essentially, skeletons have a very succinct and highly discriminative representation of the actions.  \cite{eigenjointsJournal} utilized them to extract ``eigenjoints'' for action classification using a na\"{i}ve-bayes-nearest-neighbor classifier. In \cite{HOJ3D} spherical histograms of 3D locations of the joints went through HMM to model the temporal changes and final action classification. Presence of noise in depth maps and occlusion of body parts bounds the reliability of this type of features. Another major deficiency of skeleton data is their incapacity to represent the interactions of the body with other objects which is crucial for activity interpretation.


The other group, consists of features which are extracted directly from depth maps. Most of the features in this class consider depth maps as spatio-temporal signals and tried to extract local or holistic descriptions from input sequences. \cite{msraction3ddataset} proposed a depth based action graph model in which each node indicates a salient posture and actions were represented as paths through graph nodes. To deal with occlusion and noise issues in depth maps, \cite{wangECCV12robust} proposed ``random occupancy pattern'' features and applied an elastic-net regularization \cite{elasticnet} to find the most discriminative subset of features for action recognition. STIP (space-time interest point) detection described by HOG (histogram of oriented gradients) \cite{HOG} and HOF (histogram of optical flow) was originally proposed for recognition purposes on RGB videos \cite{laptev2003spacetime}, but \cite{rgbdhudaact} showed this could be easily generalized into RGB+D signals. To improve the discrimination of descriptors, they generalized the idea of ``motion history images'' \cite{bobick2001PAMI_MHI} over depth maps. Noise-suppression could also boost up the performance of STIP detection on depth sequences \cite{xiaCVPR13spatio}. Four dimensional surface normals were shown to be very powerful representations of body movements over depth signals \cite{HON4D}. This idea was a generalization of HOG3D \cite{hog3d} into four dimensional depth videos. They quantized the 4D normal vectors of depth surfaces by taking their histograms over the vertices of a 4D regular polychoron, which were shown to be highly informative for action classification.


Regarding the strengths and weaknesses of aforementioned classes of features, we infer they are complementary to each other and to achieve higher levels of performance, we have to combine them. \cite{actionletPAMI} used histograms of 3D point clouds around the joints (LOP) to be added into skeleton based features for action classification using an ``actionlet ensemble'' framework. \cite{MMTW} added local HON4D \cite{HON4D} into joint features to learn a max\--margin temporal warping based action classifier. We utilize skeletons, LOP and HON4D as state-of-the-art depth based features to build up our multimodal input for the task of action recognition.

The main intuition behind the work of \cite{actionletPAMI} was the fact that features of few informative joints are good enough for recognizing each class of the actions. They defined ``actionlet'' as the combination of features of a limited numbers of joints and based on the discriminative power of each joint and each actionlet, they performed a data mining procedure to find the best actionlets for each class of the actions. They used mined actionlets as kernels in a multi-kernel multiclass SVM. We further extend this idea by applying group sparsity in a joint feature selection framework. To do so, we group the features of each part (joint) and applied $L^1$ norm between these groups to achieve a sparse set of active parts to represent each action class.

Mixed norms are powerful tools to inject simultaneous sparsity and coupling effects between the learning coefficients. They have been studied in a variety of fields. In statistical domain, \cite{groupLasso} proposed the ``group Lasso'', as an extension over ``Lasso'' \cite{Lasso} for a grouped variable selection in regression. \cite{zhao2009composite} introduced ``composite absolute penalty'' for hierarchical variable selection. ``Hierarchical penalization'' is also proposed to utilize prior structure of the variables for a better fitting model \cite{NIPS2007_3338}. In sparse regression, mixed norms have been used as regularization terms to link sparsity and persistence of variables \cite{Kowalski2009303}. A generalized shrinkage scheme was proposed by \cite{kowalski:inria-00369577} for structured sparse regression. \cite{6619242} used mixed norms as structured sparsity regularizers for heterogeneous feature fusion, and \cite{icml2013_wang13c} extended this idea for a multi-view clustering. \cite{wang2013robust} proposed a robust self-taught learning using mixed norms and \cite{Wang_2013_ICCV} utilized a fractional mixed norm for robust adaptive dictionary learning. In this paper, to regularize the multimodal features of each part, we apply a mixed $L^2/L^4$ norm. To achieve the sparsity between parts, we generalize this into an $L^1/L^2/L^4$ hierarchical norm.

If multiple learning tasks at hand share some inherent constituents or structures, ``Multitask Learning'' \cite{multitask1997} techniques could be globally beneficial. In joint sparse regression, multitask learning is formulated by a mixed norm. \cite{icml2009_LiuPZ09} proposed an $L^1/L^\infty$ norm to add this into Lasso for variable selection. In joint feature selection, $L^1/L^2$ norm can provide multitask learning by applying selection between the $L^2$ regularized parameters of each feature \cite{obozinski2010joint}. Same is used in \cite{multitaskTIP} as a generalization of $L^1$ norm in a multitask joint sparsity representation model to fuse complementary visual features across recognition tasks.\cite{6247908} studied different mixed norms when they applied multitask sparse learning in visual tracking and based on their experimental results, they showed $L^1/L^2$ is superior among them. In this work, we use a similar norm to utilize the shared latent factors between different binary action classifiers. We apply $L^2$ regularization over the weights corresponding to each feature across all the tasks, followed by an $L^1$ between all the features at hand.


\section{Multimodal Multipart Learning}

\subsection*{Notations}

Throughout this paper, we use bold uppercase letters to represent matrices and bold lowercase letters to indicate vectors. For a matrix $\bf X$, we denote its $j$-th row as ${\bf x}^j$ and its $i$-th column as ${\bf x}_i$.

Assume the partition $\xi$ is defined over a vector ${\bf z}$ to divide its elements into $|\xi|$ disjoint sets. We use $\xi_i$ to represent the indices of $i$-th set in $\xi$, and its corresponding elements in $\bf z$ are referred to as $\bf z^{\xi_i}$, also $z^{{\bf \xi_i},k}$ represents the $k$-th element of $\bf z^{\bf \xi_i}$. The $L^p/L^q$ norm of $\bf z$ regarding $\xi$ is represented by $\|{\bf z}\|_{q,p|\xi}$ and is defined as the $L^q$ norms of the elements inside each set of $\xi$ followed by an $L^p$ norm of the $L^q$ values across the sets; mathematically:
\begin{equation}
\label{eqn_mixednorm2}\|{\bf z}\|_{q,p|\xi}=
\left(\sum_{i=1}^{|\xi|}\|{\bf z^{\xi_i}}\|_q^p\right)^{1/p}=
\left(\sum_{i=1}^{|\xi|}{\left(\sum_{k=1}^{|\xi_i|}|z^{{\bf \xi_i},k}|^q\right)}^{p/q}\right)^{1/p}
\end{equation}
in which $|\xi_i|$ indicates the cardinality of set $\xi_i$.

Now consider the elements of each set $\xi_i$ are further partitioned by operator $\rho$ into $|\rho|$ disjoint subsets. Similarly, we indicate $j$-th $\rho$-subset of $i$-th $\xi$-set of $\bf z$ as ${\bf z}^{\bf \xi_i,\rho_j}$ and ${z}^{{\bf \xi_i,\rho_j},k}$ represents its $k$-th element. The $L^p/L^q/L^r$ norm of $\bf z$ regarding $\xi$ and $\rho$ is also represented by $\|{\bf z}\|_{r,q,p|\rho,\xi}$ and is defined as the $L^q/L^r$ norms (regarding $\rho$) of all $|\xi|$ sets followed by an $L^p$ norm of the $L^q/L^r$ values across the sets of $\xi$; mathematically:
\begin{eqnarray}
\label{eqn_mixednorm3}\|{\bf z}\|_{r,q,p|\rho,\xi}&=&
\left(\sum_{i=1}^{|\xi|}\|{\bf z^{\xi_i}}\|_{r,q|\rho}^p\right)^{1/p} \nonumber\\
&=&\left(\sum_{i=1}^{|\xi|}
\left(\sum_{j=1}^{|\rho|}{\left(\sum_{k=1}^{|\rho_j|}|z^{{\bf \xi_i,\rho_j},k}|^r\right)}^{q/r}\right)^{p/q}\right)^{1/p}
\end{eqnarray}

This representation can be easily extended into higher orders of structural mixed norms by further partitioning the subsets.

\subsection{Multipart Learning by Structured Sparsity}
\label{sec:MP}
\begin{figure}
\centering
\includegraphics[width=1\linewidth]{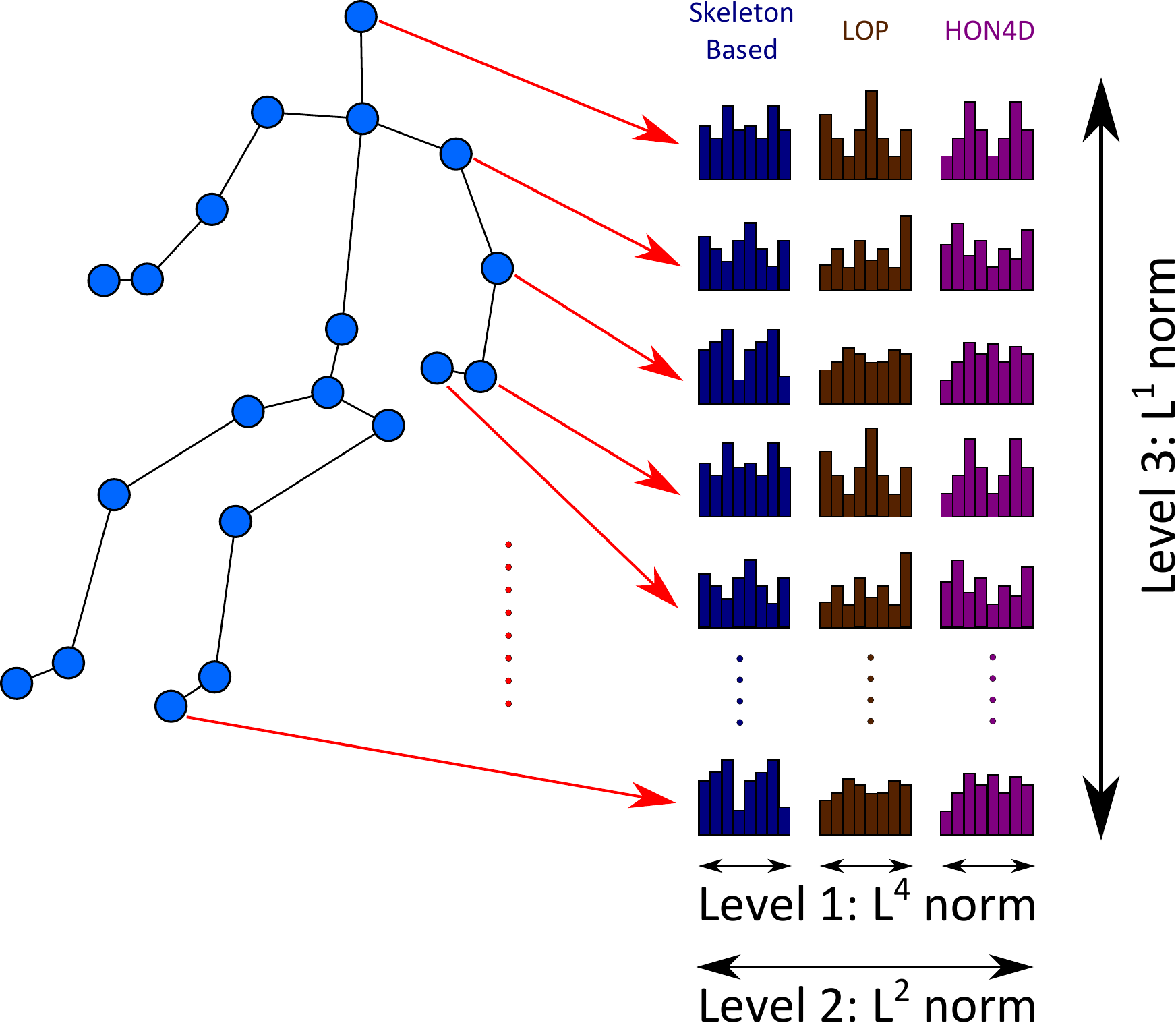}
\caption{Three Levels of the Proposed Hierarchical Mixed Norm for Multimodal Multipart Learning. We combine two levels of regularization inside modality groups and between them for each part, followed by a sparsity inducing norm between the parts to apply part selection.}
\label{fig:FrameworkHieararchical}
\end{figure}

Our purpose of learning is to recognize the actions in depth videos, based on depth based and skeleton based features extracted. The set of input features we use to describe each action sample is a combination of multimodal multipart features. The entire body is separated into a number of parts (as illustrated in \figurename{\ref{fig:FrameworkHieararchical}}) and for each part we have different types of features to represent the movement and local depth appearance. Therefore, our input feature set for each input sample, can be represented by a vector: ${\bf z} \in\mathbb{R}^d$, which consists of feature groups of different parts and modalities. Assume operator $\pi$ is partitioning $\bf z$ into $P$ parts, and $\mu$ is defined over sets of $\pi$ to further partition them based on $M$ number of features modalities. So, the hierarchy of features inside this vector is indicated by: ${\bf z} = [{{\bf z}^{\bf\pi_1}}^T,...,{{\bf z}^{\bf\pi_{P}}}^T]^T$, in which each ${\bf z}^{\bf \pi_i} = [{{\bf z}^{\bf \mu_1,\pi_i}}^T,...,{{\bf z}^{\bf \mu_{M},\pi_i}}^T]^T$.

Now the problem of multiclass action recognition can be considered as multiple binary regression based classification problems in a one versus all manner. Given $n$ training samples ${\bf X}=[{\bf x}_1,...,{\bf x}_n]$ in which ${\bf x}_i\in \mathbb{R}^{d}$ and their corresponding labels for $C$ distinct classes: ${\bf Y}=[{\bf y}_1,...,{\bf y}_C]$ with ${\bf y}_c\in{\lbrace0,1\rbrace}^n$ and $\forall i: \mathop\sum_{c=1}^{C}y_c^i=1$; we are looking for a projection matrix ${\bf W}^*\in\mathbb{R}^{d \times C}$ which minimizes a set of loss functions $J^c(\langle{\bf x}_i,{\bf w}_c^*\rangle,y_c^i)$ for all classes $c\in\lbrace1,...,C\rbrace$ and samples $i\in\lbrace1,...,n\rbrace$. Our choice for the total loss function, without loss of generality, is sum of squared errors $(\forall c: J^c(a,b) = (a-b)^2)$.

The most common shrinkage methods to regularize the learning weights against overfitting are to penalize $L^p$ norms of the learning weights for each class:
\begin{equation}
\label{eqn_regularization}{\bf w}_c^*=\mathop{\arg\!\min}_{{\bf w}_c}\sum_{i=1}^{n}J^c(\langle{\bf x}_i,{\bf w}_c\rangle,y_c^i)+\lambda \|{\bf w}_c\|_p
\end{equation}
in which $\lambda$ is the regularization factor. Employing $L^2$ norm $(p=2)$ leads into a general weight decay and minimization of the magnitude of $\bf W$, and applying $L^1$ norm $(p=1)$ yields simultaneous shrinkage and sparsity among the individual features. Such methods simply ignore the structural information between the features, which can be useful for classification; therefore, it is beneficial to embed these feature relations into our learning scheme via structured sparsity inducing mixed norms. 

In the context of depth based action recognition, features are naturally partitioned into parts. ``Actionlet ensemble'' method \cite{actionletPAMI} tried to discover discriminative joint groups using a data mining process, which led into an interesting improvement on the performance; however, their heuristic selection process is discrete and separated from the following learning step. To address these issues, we propose to apply group sparsity to perform part selection and classification in a regression based framework, in contrast to the mining based joint group discovery of \cite{actionletPAMI}.

We know that the discriminative strength of features in each part are highly correlated regarding all the classes at hand. So we expect the corresponding learning parameters (elements of each ${\bf w}_c$) to be triggered or halted concurrently within each set of $\pi$ partitioning (for each action class). To apply a grouping effect on these features, we consider each set in $\pi$ as a unit and measure its strength with an $L^2$ norm of the included learning weights. On the other hand, we seek a sparse set of parts to be activated for each class at hand, so we apply an $L^1$ norm between the $L^2$ values of the groups. Such an intuition can be formulated by an $L^1/L^2$ mixed norm based on $\pi$ for each class:
\begin{equation}
\label{eqn_singleaction}{\bf w}_c^*=\mathop{\arg\!\min}_{{\bf w}_c}\sum_{i=1}^{n}J^c(\langle{\bf x}_i,{\bf w}_c\rangle,y_c^i)+\lambda \|{\bf w}_c\|_{2,1|\pi}
\end{equation}
Adding this up for all the action classes with the same regularization factor, we have:
\begin{eqnarray}
\label{eqn_multiaction}{\bf W}^*&=&\mathop{\arg\!\min}_{\bf W}{\sum_{c=1}^{C}\sum_{i=1}^{n}J^c(\langle{\bf x}_i,{\bf w}_c\rangle,y_c^i)+\lambda \sum_{c=1}^{C}\|{\bf w}_c\|_{2,1|\pi}}\nonumber\\
\label{eqn_structuresparsity}&=&\mathop{\arg\!\min}_{\bf W}{\bf J}({\bf X}^T{\bf W},{\bf Y})+\lambda\|vec({\bf W})\|_{2,1,1|\pi,\tau}
\end{eqnarray}
in which $vec(.)$ is the vectorization operator and $\tau$ is the partitioning operator of $vec({\bf W})$ elements based on their corresponding tasks (or columns here): $\forall (k,c):\hspace{2pt}\tau(w_c^k)=c$. We will refer to this multipart learning method as ``MP''.

Minimization of (\ref{eqn_structuresparsity}) applies the desired grouping effect into the features of each part and guarantees the sparsity on the number of active parts for each class in a smooth and simpler way, compared to the actionlet method.

\subsection{Multimodal Multipart Learning via Hierarchical Mixed Norm}

In the above formulation, we apply an $L^2$ regularization norm over heterogeneous features of all the modalities for each part, and ignore the modality structures between them. In other words, applying a general $L^2$ norm may cause the suppression of the information at some dimensions. These issues are more severe when training samples are limited (which is the case for action recognition in depth), in which it might lead to weak generalization of the learning. 

To overcome these limitations, we utilize $L^\infty$ to regularize the coefficients inside each modality, so that ``diversity'' \cite{Kowalski2009303} can be encouraged. It is already known that the behavior of $L^p$ norm for $p>2$ rapidly moves towards $L^\infty$ \cite{5948411}; since $L^\infty$ is not easy to optimize directly, we picked $L^4$ as the most efficient approximation of it. Higher order norms like $L^6$ apply the same effect but with a slightly more expensive processing cost. 


By applying the $L^4$ norm to regularize the weights in each modality group of each part, now we have a three-level $L^1/L^2/L^4$ mixed norm. Inner $L^4$ gives more ``diversity''  to regularize the features inside each partiality-modality subset. $L^2$ norm employs a magnitude based regularization over the $L^4$ values to link different modalities of each part, and the outer $L^1$ applies the soft part selection between the $L^2/L^4$ values of each action class (\figurename{\ref{fig:FrameworkHieararchical}}).

Replacing the previous structured norm by the proposed hierarchical mixed norm in (\ref{eqn_structuresparsity}), we have:
\begin{eqnarray}
\label{eqn_structuresparsity421S}{\bf W}^*&=&\mathop{\arg\!\min}_{\bf W}{\bf J}({\bf X}^T{\bf W},{\bf Y})+\lambda\sum_{c=1}^{C}\|{\bf w}_c\|_{4,2,1|\mu,\pi}\nonumber\\
\label{eqn_structuresparsity421}&=&\mathop{\arg\!\min}_{\bf W}{\bf J}({\bf X}^T{\bf W},{\bf Y})+\lambda\|vec({\bf W})\|_{4,2,1,1|\mu,\pi,\tau}
\end{eqnarray}
here, $\pi$ indicates the partitioning of features based on their source body part, and $\mu$ represents further partitioning of each part's set regarding the modalities of the features. In the rest of this paper, we use the abbreviation ``MMMP'' to refer to this method. It is worthwhile to note, changing the inner norm to $L^2$ will reduce the hierarchical norm into a two level mixed norm, i.e.  $\|vec({\bf W})\|_{2,2,1,1|\mu,\pi,\tau}=\|vec({\bf W})\|_{2,1,1|\pi,\tau}$ derived directly from the definition of hierarchical norm (\ref{eqn_mixednorm3}).

When different learning tasks have similar latent features, ``Multitask Learning'' \cite{multitask1997} techniques can improve the performance of the entire system by applying information sharing between the tasks. Here we are learning classifiers for $C$ different classes which essentially have lots of latent components in common, so pushing them to share some features is beneficial for the classification task. This can be done by applying an $L^2$ grouping on all the weights corresponding to each individual feature. Each of these $L^2$ values represents the magnitude of strength for its corresponding feature among all the tasks. Then applying an $L^1$ over the magnitudes can apply a shared variable selection considering all the tasks. Adding the new multitask term into (\ref{eqn_structuresparsity421}), we have:
\begin{eqnarray}
\label{eqn_multitaskl1l2}{\bf W}^*=\mathop{\arg\!\min}_{\bf W}{\bf J}({\bf X}^T{\bf W},{\bf Y})+\lambda_1\sum_{k=1}^{d}\|{\bf w}^k\|_2\nonumber\\
+\lambda_2\|vec({\bf W})\|_{4,2,1,1|\mu,\pi,\tau}\\
\label{eqn_multitaskl21}=\mathop{\arg\!\min}_{\bf W}{\bf J}({\bf X}^T{\bf W},{\bf Y})+\lambda_1\|{vec(\bf W)}\|_{2,1|\phi}\nonumber\\
+\lambda_2\|vec({\bf W})\|_{4,2,1,1|\mu,\pi,\tau}
\end{eqnarray}
here, $d$ is the number of rows in $\bf W$ which is equal to the size of the entire feature vector, and $\phi$ defines the partitioning of $vec({\bf W})$ elements based on their corresponding individual features: 
$\forall (k,c):\hspace{2pt}\phi(w_c^k)=k$.

Combining these two regularization terms can be considered as a trade off between sparsity and persistence of features \cite{kowalski09sparsity} based on their relations across the parts, modalities, and between the action classes.

In our experiments, we use $P=20$ body joints as partitioning operator $\pi$. Since each column of $\bf W$ has the same hierarchical partitioning as input features: ${\bf W}=[{\bf w}_c^j]$, in which $c$ counts the number of classes and $j$ counts the feature groups for $P$ joints. The features for each joint come from ${\bf M}=3$ different modalities: skeletons, LOP, and HON4D; this defines the $\mu$ operator. Therefore, each ${\bf w}_c^j=[{{\bf w}_c^{j,1}}^T,...,{{\bf w}_c^{j,M}}^T]^T$, in which each ${{\bf w}_c^{j,m}}$ is the corresponding weight elements to class $c$, joint $j$ and modality $m$. This way (\ref{eqn_multitaskl21}) will be expanded to:
\begin{eqnarray}
\label{eqn_expanded}{\bf W}^*&=&\mathop{\arg\!\min}_{\bf W}\|{\bf X}^T{\bf W}-{\bf Y}\|_F^2+\lambda_1\sum_{k=1}^{d}\|{\bf w}^k\|_2\nonumber\\
&+&\lambda_2\sum_{c=1}^{C}\sum_{j=1}^{P}({{\sum_{m=1}^{M}}\|{\bf w}_c^{j,m}\|_{4}^2})^{1/2}
\end{eqnarray}

\subsection{Two Step Learning Approach}
The downside of current formulation is the large number of weights to be learned simultaneously, compared to the size of training samples which are highly limited in current depth based action recognition benchmarks. To resolve this, we first learn the partially optimum weights for multipart features of each modality separately and then fine-tune them by the proposed multimodal multipart learning. 

To learn the partially optimum weights for each modality $m$, we optimize:
\begin{eqnarray}
\label{eqn_warmstartpoint}\widehat{\bf W}_{\bf m}=\mathop{\arg\!\min}_{\bf W_{\bf m}}{\bf J}({\bf X}_{\bf m}^T{\bf W}_{\bf m},{\bf Y})+\hat{\lambda}_1\|{vec(\bf W_{\bf m}})\|_{2,1|\phi}\nonumber\\
+\hat{\lambda}_2\|vec({\bf W}_{\bf m})\|_{2,1,1|\pi,\tau}
\end{eqnarray}

After achieving the partially optimum point for each modality, we merge the $\widehat{\bf W}_{\bf m}$ values for all $M$ modalities:
\begin{eqnarray}
\widehat{{\bf W}}=\lbrack\widehat{{\bf W}}_1^T,...,\widehat{{\bf W}}_M^T\rbrack^T
\end{eqnarray}

Next is to fine-tune the weights in the multimodal-multipart learning fashion, on a neighborhood of $\widehat{\bf W}$ values. To do so, we expect the global optimum weight not to diverge too much from their partially optimal points:
\begin{eqnarray}
\label{eqn_structuresparsity421WR}{\bf W}^*=\mathop{\arg\!\min}_{\bf W}{\bf J}({\bf X}^T{\bf W},{\bf Y})+\lambda_1\|{vec(\bf W})\|_{2,1|\phi}\nonumber\\
+\lambda_2\|vec({\bf W})\|_{4,2,1,1|\mu,\pi,\tau}+\lambda_3\|{\bf W}-{\widehat{\bf W}}\|_F^2
\end{eqnarray}

The last term in (\ref{eqn_structuresparsity421WR}) will limit the deviation of learning weights from their partially optimal point, as we expect them to be just fine-tuned in this step.

Upon optimization over training data, the detection of the learned classifier for each testing sample ${\bf x}_i$ can be obtained by:
\begin{eqnarray}
\label{eqn_detectiontest}f({\bf x}_i)=\mathop{\arg\!\max}_{c}{\langle{\bf x}_i,{\bf w}_c^*\rangle}
\end{eqnarray}
The optimization steps are all done by ``L-BFGS'' algorithm using off-the-shelf ``minFinc'' tool \cite{schmidt2005minfunc}.

\section{Experiments}
This section describes our experimental setup details and then provides the results of the proposed method on three depth based action recognition benchmarks.

\subsection{Experimental Setup}
All the provided experiments are done on Kinect based datasets. Kinect captures RGB frames, depth map signals and 3D locations of major joints. To have a fair comparison with other depth based methods, we ignore the RGB signals. Skeleton extraction is done automatically by Kinect's SDK based on the part-based human pose recognition system of \cite{shotton2011CVPR}. On each frame, we have an estimation of 3D positions of 20 joints in the body. All of our features are defined based on these joints as the multipart partitioning operator ($\pi$); therefore, each feature necessarily belongs to one of these parts.

To represent skeleton based features, first we normalize the 3D locations of joints against size, position and direction of the body in the scene. This normalization step eases the task of comparison between body poses. On the other hand, the extracted body locations and directions could also be highly discriminative for some action classes like ``walking'' or ``lying down''; therefore we add them into the features under a new auxiliary part. To encode the dynamics of skeleton based features, we apply ``Fourier temporal pyramid'' as suggested by \cite{actionletPAMI} and keep first four frequency coefficients of each short time Fourier transformation. This leads into a feature vector of size 1,876 for each action sample.

In addition to skeleton based features, other modalities we use are local HON4D \cite{HON4D} and LOP \cite{actionletPAMI} to represent depth based local dynamics and appearance around each joint. On each frame, LOPs are extracted on a (96,96,320)\--sized depth neighborhood of each joint, which is divided into $3\times3\times4$ number of (32,32,80)\--sized bins. To represent LOP based kinetics, we use a similar Fourier temporal pyramid transformation. HON4D features are also extracted locally over the location of joints on each frame. We encode HON4D features using LLC (locality-constrained linear coding) \cite{LLC} to reduce their dimensionality while preserving the locality of 4D surface normals. Dictionary size of 100 is picked for the clustering step. LLC codes go through a max pooling over a 3 level temporal pyramid. Dimension of the features for LOP and HON4D are 5,040 and 14,000 respectively. The overall dimensionality of input features for each sample is 20,916.

\subsection{MSR-DailyActivity3D Dataset}

\begin{table}[!t]
	\renewcommand{\arraystretch}{1.3}
	\caption{Subject-wise Cross-Validation Performance Comparison of the Proposed Hierarchical Mixed Norm with Plain and Multipart Group Sparsity Norm on the MSR-DailyActivity Dataset}
	\label{table_msrdailyactivitynorms}
	\centering
	\begin{tabular}{|c||c||c|}
		\hline
		\bfseries Method & \bfseries Structure/Hierarchical Norm Used & \bfseries Accuracy \\\hline
		\hline
		$L^2$ & $\|vec({\bf W})\|_2^2$ & 80.61$\pm$2.49\%\\\hline
		MP & $\|vec({\bf W})\|_{2,1,1|\pi,\tau}$ & 81.55$\pm$2.43\%\\\hline
		MMMP & $\|vec({\bf W})\|_{4,2,1,1|\mu,\pi,\tau}$ & 84.03$\pm$2.16\%\\\hline
	\end{tabular}
\end{table}

\begin{table}[!t]
	\renewcommand{\arraystretch}{1.3}
	\caption{Performance Comparison of the Proposed Method Using Plain/Structured/Hierarchical Norms on the Standard Evaluation Split of the MSR-DailyActivity Dataset}
	\label{table_msrdailyactivitysettings}
	\centering
	\begin{tabular}{|c||c||c|}
		\hline
		\bfseries Method & \bfseries Structure/Hierarchical Norm Used & \bfseries Accuracy \\\hline
		\hline
		$L^1$ & $\|vec({\bf W})\|_1$ & 86.88\%\\\hline
		$L^2$ & $\|vec({\bf W})\|_2^2$ & 87.50\%\\\hline
		MP & $\|vec({\bf W})\|_{2,1,1|\pi,\tau}$ & 88.13\%\\\hline
		MMMP & $\|vec({\bf W})\|_{4,2,1,1|\mu,\pi,\tau}$ & 91.25\%\\\hline
	\end{tabular}
\end{table}

\begin{table}[!t]
	\renewcommand{\arraystretch}{1.3}
	\caption{Performance Comparison on the Standard Evaluation Split of the MSR-DailyActivity Dataset using Single Modality and Multimodal Features.}
	\label{table_msrdailyactivity}
	\centering
	\begin{tabular}{|c|c||c|}
		\hline
		\bfseries Method & \bfseries Modalities & \bfseries Accuracy
		\\\hline\hline
		Actionlet Ensemble \cite{actionletPAMI} & LOP & 61\%\\
		\hline
		{\bf Proposed MP} & LOP &{\bf 79.38\%}\\
		\hline\hline
		Orderlet Mining \cite{Orderlet} & Skeleton & 73.8\%\\
		\hline
		Actionlet Ensemble \cite{actionletPAMI} & Skeleton & 74\%\\
		\hline
		{\bf Proposed MP} & Skeleton & {\bf 79.38\%}\\
		\hline\hline
		Local HON4D \cite{HON4D} & HON4D & 80.00\%\\
		\hline
		{\bf Proposed MP} & HON4D &{\bf 81.88\%}\\
		\hline\hline
		Actionlet Ensemble \cite{actionletPAMI} & Skeleton+LOP & 85.75\%\\
		\hline
		{\bf Proposed MMMP} & Skeleton+LOP & {\bf 88.13\%}\\
		\hline\hline
		MMTW \cite{MMTW} & Skeleton+HON4D & 88.75\%\\
		\hline
		{\bf Proposed MMMP}& Skeleton+HON4D & {\bf 89.38\% }\\
		\hline\hline
		DSTIP \cite{xiaCVPR13spatio} & DCSF+LOP & 88.20\%\\
		\hline\hline
		{\bf Proposed MMMP}& Skeleton+LOP+HON4D & {\bf 91.25\% }\\
		\hline
	\end{tabular}
\end{table}

According to its intra-class variations and choices of action classes, MSR-DailyActivity dataset \cite{actionletPAMI}, is one of the most challenging benchmarks for action recognition in depth sequences. It contains RGB, depth, and skeleton information of 320 action samples, from 16 classes of daily activities in a living room. Each activity is done by 10 distinct subjects in two different ways and evaluations are applied over a fixed cross-subject setting; first five subjects are taken for training and others for testing. Unlike other datasets, MSR-DailyActivity has a more realistic variation within each class. Subjects used both hands randomly to do the activities, and samples of each class are captured in different poses.

First, to verify the strengths of our proposed hierarchical mixed norm, we evaluate the performance of the classification in a subject-wise cross-validation scenario. We evaluate the performance of the plain $L^2$ norm, the multipart structured norm (MP), and the proposed hierarchical mixed norm (MMMP), in all 252 possible train/test splits of 5 out of 10 subjects. To have a proper comparison between these norms, we have not applied the multitask term. The results of this experiment are shown in Table \ref{table_msrdailyactivitynorms}. Adding part based grouping, when it ignores the modality associations between the features, can slightly improve the performance from 80.61\% into 81.55\%. By adding multimodality grouping and applying the proposed hierarchical mixed norm, improvement is more significant and reaches 84.03\%.

Next, we verify the results of our method by applying mentioned norms on the standard train/test split of the subjects. As provided in Table \ref{table_msrdailyactivitysettings}, applying simple feature selection using a plain $L^1$ norm leads into 86.88\% of accuracy. By applying a plain $L^2$ norm on all the features we get 87.50\%. Multipart learning regardless of heterogeneity of the modalities leads into 88.13\%. Finally by adding the multipart learning via the proposed hierarchical mixed norm we reach the interesting accuracy of 91.25\% on this dataset. Applying higher orders for the inner-most norm (like $L^1/L^2/L^6$) achieved the same level of accuracy at a slightly higher processing time.

To assess the strength of the proposed multipart learning, we evaluate our method on single modality setting using (\ref{eqn_warmstartpoint}). As shown in Table \ref{table_msrdailyactivity}, on skeleton based features, we got 79.38\% compared to 74\% of the baseline actionlet method. Using LOPs, our method achieved 79.38\% which is more than 18\% higher than the actionlet's performance. For local HON4D features, we achieved 81.88\% compared to 80.00\% of the baseline local HON4D method.
Now we use the partially learned weights of single modality multipart learning and employ them for the optimization of (\ref{eqn_structuresparsity421WR}) to learn globally optimum projections. First we try the combination of skeleton based features with LOP. Using proposed learning, we get 88.13\% of accuracy which outperforms the baseline's best result of 85.75\%. \cite{MMTW} used skeleton and HON4D features in a temporal warping framework and got 88.75\%. Our method outperforms it using the same set of features by achieving 89.38\% of accuracy. And finally using all three modalities, our method achieves the performance level of 91.25\%. Table \ref{table_msrdailyactivity} shows the complete set of results for this experiment. 

Our implementation is done in MATLAB, and not fully optimized for time efficiency. The average training and testing time of MMMP on a $3.2~GHz$ Core-i5 machine are $170$ and $2\times10^{-4}$ seconds respectively, with no parallel processing.

It is worth pointing out some of the published works on this dataset applied other train/test splits, e.g. \cite{AlthloothiPR} reported 93.1\% of accuracy on a leave-one-subject-out cross validation. On this setup, proposed MMMP method achieves 97.5\%.



\subsection{MSR-Action3D Dataset}
\begin{table}[!t]
	\renewcommand{\arraystretch}{1.3}
	\caption{Average Cross Subject Performance for MSR-Action3D Dataset on Three Action Subsets of \cite{msraction3ddataset}}
	\label{table_msraction3dT1}
	\centering
	\begin{tabular}{|c||c|}
		\hline
		\bfseries Method (protocol of \cite{msraction3ddataset}) & \bfseries Accuracy\\
		\hline
		\hline Action Graph on Bag of 3D Points \cite{msraction3ddataset} & 74.7\%\\
		\hline Histogram of 3D Joints \cite{HOJ3D} & 79.0\%\\
		\hline EigenJoints \cite{eigenjointsJournal} & 83.3\%\\
		\hline Random Occupancy Patterns \cite{wangECCV12robust} & 86.5\%\\
		\hline Depth HOG \cite{DHOG} & 91.6\%\\
		\hline Lie Group \cite{VemulapalliCVPR14} & 92.5\%\\
		\hline JAS+HOG$^2$ \cite{hog2-ohnbar} & 94.8\%\\
		\hline DL-GSGC+TPM \cite{Luo_2013_ICCV} & 96.7\%\\
		\hline\hline
		{\bf Proposed MMMP} & {\bf 98.2\%}\\
		\hline
	\end{tabular}
\end{table}

\begin{table}[!t]
	\renewcommand{\arraystretch}{1.3}
	\caption{Performance Comparison for MSR-Action3D Dataset Over All Action Classes}
	\label{table_msraction3dT2}
	\centering
	\begin{tabular}{|c||c|}
		\hline
		\bfseries Method (protocol of \cite{actionletPAMI})& \bfseries Accuracy\\
		\hline
		\hline Depth HOG \cite{DHOG} (as reported in \cite{MMTW}) & 85.5\%\\
		\hline Actionlet Ensemble \cite{actionletPAMI} & 88.2\%\\
		\hline HON4D \cite{HON4D} & 88.9\%\\
		\hline DSTIP \cite{xiaCVPR13spatio} & 89.3\%\\
		\hline Lie Group \cite{VemulapalliCVPR14} & 89.5\%\\
		\hline HOPC \cite{HOPC} & 91.6\%\\
		\hline Max Margin Time Warping \cite{MMTW} & 92.7\%\\
		\hline 
		\hline
		{\bf Proposed MMMP} & {\bf 93.1\%}\\
		\hline
	\end{tabular}
\end{table}


MSR-Action3D \cite{msraction3ddataset} is another depth based action dataset which provided depth sequences and skeleton information of 567 samples for 20 action classes. Actions are done by 10 different subjects, two or three times each. Evaluations are applied over another fixed cross-subject setting; Odd numbered subjects are taken for training and evens for testing. On one hand, depth sequences in this dataset have clean background which eases the recognition, and on the other hand, number of classes are higher than other datasets which could be a challenge for classification.

The reported results on this dataset are divided in two different scenarios. First is the average cross subject performance on three action subsets defined in \cite{msraction3ddataset}, and second is the overall cross subject accuracy regardless of subsets, as done in \cite{actionletPAMI}. Following \cite{VemulapalliCVPR14}, we call them as protocols of \cite{msraction3ddataset} and \cite{actionletPAMI}. Tables \ref{table_msraction3dT1} and \ref{table_msraction3dT2} show the results. Although we still have the highest accuracy among the reported results, the achieved margin is not as large as other datasets. This is because of the simplicity of actions in this dataset. Since there is not any interaction with other objects, most of the classes are highly distinguishable using skeleton only features; therefore our multimodality could not boost up the results that much, but the multipart learning still shows its advantage over other methods. 


\subsection{3D Action Pairs Dataset}
To emphasize the importance of the temporal order of body poses on the meaning of the actions, \cite{HON4D} proposed 3D Action Pairs dataset. It covers 6 pairs of similar actions. The only difference between each pair is their temporal order so they have similar skeleton, poses, and object shapes. Each action is performed by 10 subjects, 3 times. First five subjects are taken for testing and others for training. Based on the fewer number of the action classes and absence of intra-class variations, this is the easiest benchmark among depth based action recognition datasets and other methods already achieved very high accuracies on it.

Here we apply our full multimodal multipart learning method using all three available modalities of features. As shown in Table \ref{table_3dactionpairs}, the proposed method, outperforms all others and saturates the benchmark by achieving the perfect performance level on this dataset.

\section{Conclusion}

This paper presents a new multimodal multipart learning approach for action classification in depth sequences. We show that a sparse combination of multimodal part-based features can effectively and discriminatively represent all the available action classes at hand. Based on the nature of the problem, we utilize a heterogeneous set of features from skeleton based 3D joint trajectories, depth occupancy patterns and histograms of depth surface normals and show the proper way of using them as multimodal features set for each part.

The proposed method does the group feature selection, weight regularization, and classifier learning in a consistent optimization step. It applies the proposed hierarchical mixed norm to model the proper structure of multimodal multipart input features by applying a diversity norm over the coefficients of each part-modality group, linking different modalities of each part by a magnitude based norm, and utilizing a soft part selection by a sparsity inducing norm.

The provided experimental evaluations on three challenging depth based action recognition datasets show the proposed method can successfully apply the structure of the input features into a concurrent group feature selection and learning and confirm the strengths of the suggested framework compared to other methods.

\begin{table}[!t]
	\renewcommand{\arraystretch}{1.3}
	\caption{Performance Comparison for 3D Action Pairs Dataset}
	\label{table_3dactionpairs}
	\centering
	\begin{tabular}{|c||c|}
		\hline
		\bfseries Method & \bfseries Accuracy\\\hline
		\hline Depth HOG \cite{DHOG} (as reported in \cite{MMTW}) & 66.11\%\\
		\hline Actionlet Ensemble \cite{actionletPAMI} (as reported in \cite{MMTW}) & 82.22\%\\
		\hline HON4D \cite{HON4D} & 96.67\%\\
		\hline Max Margin Time Warping \cite{MMTW} & 97.22\%\\
		\hline HOPC \cite{HOPC} & 98.33\%\\
		\hline\hline
		{\bf Proposed MMMP} & {\bf 100.0\%}\\
		\hline
	\end{tabular}
\end{table}



\ifCLASSOPTIONcaptionsoff
  \newpage
\fi



\bibliographystyle{IEEEtran}
\end{document}